\documentclass[runningheads]{llncs}
\usepackage{graphicx}
\usepackage{bm}
\usepackage{amsmath}
\usepackage{amsfonts, amssymb}
\usepackage{multirow}
\usepackage{booktabs}
\usepackage{subfigure}
\begin{document}
\title{MOEF: Modeling Occasion Evolution in Frequency Domain for Promotion-Aware Click-Through Rate Prediction}

\titlerunning{MOEF}

\author{
    Xiaofeng Pan\inst{1} \and
    Yibin Shen\inst{1} \and
    Jing Zhang\inst{2} \and 
    Xu He\inst{1} \and
    Yang Huang\inst{1} \and
    Hong Wen\inst{1} \and
    Chengjun Mao\inst{1} \and
    Bo Cao\inst{1}
}
\authorrunning{X. Pan et al.}
%
\institute{
    Alibaba Group, Hangzhou, China \\
    \email{pxfvintage@163.com, \{shilou.syb, hx234688, hy234680, qinggan.wh, chengjun.mcj, zhizhao.cb\}@alibaba-inc.com} 
    \and
    The University of Sydney \\
    \email{jing.zhang1@sydney.edu.au} 
}

\maketitle              

\begin{abstract}
Promotions are becoming more important and prevalent in e-commerce to attract customers and boost sales, leading to frequent changes of occasions, which drives users to behave differently. In such situations, most existing Click-Through Rate (CTR) models can't generalize well to online serving due to distribution uncertainty of the upcoming occasion. In this paper, we propose a novel CTR model named MOEF for recommendations under frequent changes of occasions. Firstly, we design a time series that consists of occasion signals generated from the online business scenario. Since occasion signals are more discriminative in the frequency domain, we apply Fourier Transformation to sliding time windows upon the time series, obtaining a sequence of frequency spectrum which is then processed by Occasion Evolution Layer (OEL). In this way, a high-order occasion representation can be learned to handle the online distribution uncertainty. Moreover, we adopt multiple experts to learn feature representations from multiple aspects, which are guided by the occasion representation via an attention mechanism. Accordingly, a mixture of feature representations is obtained adaptively for different occasions to predict the final CTR. Experimental results on real-world datasets validate the superiority of MOEF and online A/B tests also show MOEF outperforms representative CTR models significantly.

\end{abstract}
\section{Introduction} 
Click-Through Rate (CTR) prediction has been one of the most important tasks in recommender systems \cite{sarwar2001item,wen2020entire,zhang2020empowering} since it is directly related to user satisfaction, efficiency, and revenue. With the rapid progress of deep neural models, most of CTR models use high-order interactions of features to improve their representation ability \cite{cheng2016wide,guo2017deepfm,qu2018product,liu2020autofis,zhu2021aim}, and leverage sequential user behaviors to model users in a dynamic manner \cite{zhou2018deep,zhou2019deep,ren2019lifelong,xu2020deep,xiao2020deep}.

\begin{figure*}[t]
    \centering
    \includegraphics[width=1.0\linewidth]{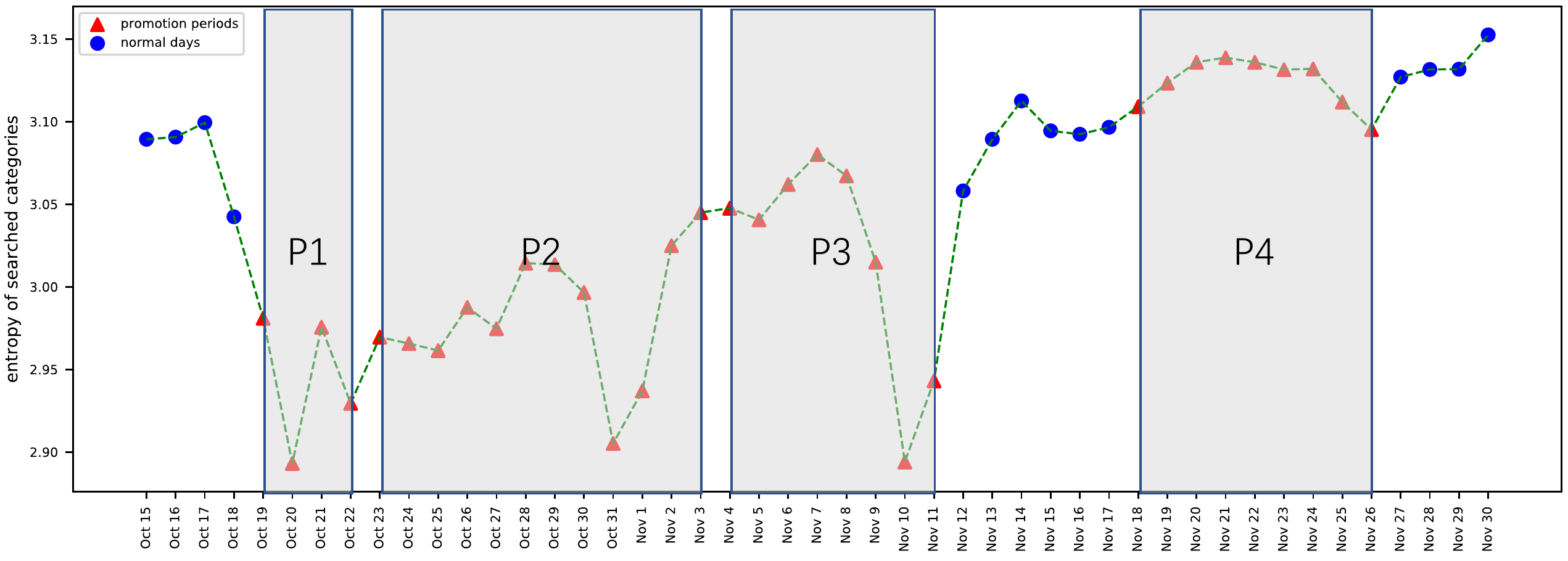}
    \caption{The entropy of searched categories is calculated as $H=-\sum\,p_i\,log\,p_i$, where $p_i$ denotes the global search ratio of the $i$-th category. $H$ intuitively shows the diversity of user interests and its change reflects the change of user data distribution.}
    \label{fig:distribution}
\end{figure*}

Most previous works assume that users' preferences are coherent and change smoothly over time. However, users' behaviors can be significantly influenced by different occasions \cite{wang2020time} which are fine-grained time periods related to particular times or events. Especially with the intensification of e-commerce competition, online promotions are becoming more frequent and diverse, which differ from each other in many factors, $e.g.$, scale, duration, density, and regularity in time, resulting in frequent changes of occasions. From Figure~\ref{fig:distribution}, we can observe that the entropy of searched categories, an indicator of user data distribution, fluctuated dramatically as four promotions occurred. The fluctuations during different promotions differed greatly and shared no regularity, illustrating that there exists considerable distribution uncertainty under frequent changes of occasions, although the schedule of promotions is available beforehand. Besides, with different occasions included, training samples can be non-identically distributed, imposing extra difficulty on model training. To achieve promotion-aware CTR prediction
, we need to tackle these two issues simultaneously.

An intuitive idea to deal with recommendations for e-commerce promotions 
is leveraging years of plenty of historical data to train models with better generalization performance. However, directly introducing data with a large time span results in discrepant data distributions between training and serving, which degrades the generalization performance \cite{pan2022metacvr}. Besides, promotions driven by emerging consumption trends are inaccessible in historical data. As far as we know, the most widely adopted way to handle recommendations for promotions is temporarily customizing or adding extra models, which highly relies on expert experience to decide when to switch models for online serving because it's uncertain when data distribution will change.

Recently, several studies, $e.g.$, STAR \cite{sheng2021one}, TREEMS \cite{niu2021heterogeneous}, and SAR-Net \cite{shen2021sar}, have been carried out to handle Multi-Scenario CTR prediction, which seems applicable to our case. 
In general, these methods leverage scenario features to learn data distributions of different scenarios to facilitate the Mixture-of-Experts \cite{jacobs1991adaptive} so they can serve all scenarios with one model. Each recommendation scenario locates spatial differently and can be treated as stable over time, with explicit features related to data distribution ($i.e.$, scenario features such as scenario id). In contrast, occasions intertwine with each other in time, and no explicit features are available to provide certainty of data distribution and distinguish different occasions, so existing Multi-Scenario CTR models don't apply to our case. Another line of works \cite{wang2020time,wang2020next,wang2021session,li2020deep} tries to make models more sensitive to time. In \cite{wang2020time}, a model is developed to adapt to different occasions by learning representations indexed by timestamps. However, occasions resulting from promotions are not necessarily related to timestamps. In \cite{li2020deep}, item behaviors \cite{du2016recurrent}, $i.e.$, a set of users who interact with this item, are introduced via a time-sensitive neural structure, which models items in a dynamic manner to strengthen the ability to predict users' emerging interests. The methods proposed in \cite{wang2020next,wang2021session} follow the similar idea of modeling items in a dynamic manner by adopting hypergraphs. However, local changes captured from item behaviors are not equivalent to global changes of data distribution.

Based on these observations, we propose a novel CTR model named MOEF which models occasion evolution to obtain a well-learned occasion representation to modulate the learning of user-item feature representations. Specifically, with inspiration from time series prediction \cite{crone2010feature,yue2022ts2vec}, we generate occasion signals (detailed in Section~\ref{sec:preliminary}) from our online business scenario with a proper sampling interval, constructing a time series which is then processed by an elaborated Occasion Representation Network (ORN). In the ORN, we obtain a sequence of frequency spectrum by applying Fast Fourier Transformation (FFT) \cite{soliman1990continuous} to time windows that slide on the time series with proper window size and stride, since occasion signals are more discriminative in the frequency domain. Then we process the sequence of frequency spectrum by Occasion Evolution Layer (OEL), where deep neural structures for sequential modeling such as LSTM \cite{hochreiter1997long} and Transformer-Encoder \cite{vaswani2017attention} can be used to model occasion evolution. In this way, a high-order occasion representation can be learned to help tackle the online distribution uncertainty. Meanwhile, multiple experts are adopted to learn feature representations, mitigating the mutual interference between non-identically distributed training data. Under modulation of the occasion representation, multiple experts can learn feature representations from different aspects. At last, a mixture of experts is calculated via an attention mechanism and used for the final prediction. Our main contributions are summarized as follows:
\begin{itemize} 
    \item To the best of our knowledge, this is the first study of CTR prediction concerning e-commerce promotions. We introduce occasion signals from a perspective of time series prediction and propose a novel MOEF model to achieve promotion-aware CTR prediction.
    \item We model occasion evolution in the frequency domain to facilitate learning a good occasion representation, which is further used to modulate the learning of multiple experts via an attention mechanism. In this way, both the distribution uncertainty and training difficulty resulting from non-identically distributed training data are handled.
    \item Experiments on real-world datasets and online tests demonstrate the superiority of MOEF over representative methods. We also perform extensive analyses for better interpretability. The code is publicly available\footnote{https://github.com/AaronPanXiaoFeng/MOEF}.
\end{itemize}

\section{Related Work}
\subsection{CTR Prediction}
CTR prediction has become a crucial part of many online applications, such as search engines, recommender systems, and online advertising. To better capture high-order feature interactions, there are some representative DNN-based CTR models have been proposed, including Wide\&Deep \cite{cheng2016wide}, PNN \cite{qu2016product}, DeepFM \cite{guo2017deepfm}, and PIN \cite{qu2018product}. Combining the idea of AutoML, AutoFIS \cite{liu2020autofis} proposes to find useful feature interactions, and AIM\cite{zhu2021aim} further decides how the interaction should be modeled. Meanwhile, since the sequence of user behaviors contains rich information of users' interests and preferences, there is increasing attention on sequential modeling in online systems. Given a target item, DIN \cite{zhou2018deep} introduces the attention mechanism to activate the historical behaviors. To capture dependencies between sequential behaviors, DIEN \cite{zhou2019deep} adopts a two-layer RNN structure to model the evolving process of specific interest for different target items. MIMN \cite{pi2019practice} proposes a memory-based architecture to aggregate features and tackle the challenge of long-term user interest modeling. DMIN \cite{xiao2020deep} models user’s multiple interests by a special designed extractor layer. For better modeling of the temporal effects, HPMN \cite{ren2019lifelong} is proposed to capture the periodic patterns of user interests with a hierarchical recurrent memory network, while TIEN \cite{li2020deep} models items in a dynamic manner by using item behaviors to strengthen the ability to predict users' emerging interests. However, none of these models pay enough attention to frequent changes of occasions resulting from promotions. More recently, Multi-Scenario CTR methods \cite{sheng2021one,niu2021heterogeneous,shen2021sar} attempt to serve different scenarios with a unified model, which however are not able to deal with promotions as discussed in the introduction.

\subsection{Mixture-of-Experts}
For deep learning models, ensemble of sub-networks has been proven to be effective in improving model performance \cite{hinton2015distilling}. The Mixture-of-Experts (MoE) \cite{jacobs1991adaptive} first proposes to share some experts and combine them based on a gating network, which is very helpful when dealing with complex problems that may contain many sub-problems, each requiring different experts. Therefore, MoE has been widely used as basic blocks in recent deep models \cite{eigen2013learning,shazeer2017outrageously}. Further, MMoE \cite{ma2018modeling} extends MoE with multiple gates to model task differences, and PLE \cite{tang2020progressive} further improves the design of connection routing and gating mechanisms. We took inspiration from these works by using MoE to learn different feature representations with different experts.

\section{Preliminaries} \label{sec:preliminary}
\subsection{Problem Formulation} \label{subsec:problem}
In CTR prediction, the model takes input as $(\bm{x}, y) \sim (X,Y)$, where $\bm{x}$ is the feature and $y \in \{0,1\}$ is the click label. Specifically, the features in this work consist of six parts: 1) user behavior sequence $\bm{x}^{seq}$; 2) user features $\bm{x}^u$ including user profile and user statistic features; 3) item features $\bm{x}^i$ such as item id, category, brand, and related statistic features; 4) context features $\bm{x}^c$ such as position and time information; 5) occasion signals $\bm{x}^o$, $i.e.$, statistics of the online business scenario ($e.g.$, active user amount, gross merchandise volume, and amount of add-to-cart behavior) which are calculated with a proper sampling interval and cover the time frame of last few hours.

The goal of CTR prediction is to learn a model $f_{\theta}$ parameterized with $\theta$ that minimizes the generalization error:
\begin{equation}
    \theta^{*}= \underset{\theta}{min} E_{(\bm{x}, y) \sim (X,Y)}[L(\bm{x},y;f_{\theta}(\bm{x}))],
\end{equation}
where $L$ is the loss function. In the situation of frequent promotions, CTR prediction becomes highly challenging due to two issues: 1) data distribution of the upcoming occasion is uncertain, and 2) the distribution of training dataset is non-identical since samples are collected from different occasions.

\subsection{Occasions and Occasion Signals} \label{subsec:occasion}
Similar but slightly different to global occasions discussed in \cite{wang2020time}, occasions in this work are more fine-grained time periods when users' behaviors change significantly. Thus, occasions derived from promotions are not explicit. To observe the changes of occasions, we design a set of statistical features of the online business scenario, $i.e.$,
\begin{equation}
    \bm{x}^o=\{\bm{x}^o_1, ..., \bm{x}^o_i, ..., \bm{x}^o_M\},
\end{equation}
where $M$ denotes the number of occasion signals, and $\bm{x}^o_i=[x^o_{i1}, x^o_{i2}, ..., x^o_{iN}]$ denotes the $i$-th occasion signal which is recorded in the form of a time-domain sequence. All occasion signals are calculated with a sampling interval of $T$ and only keep the recent $N$ values. It's noteworthy that $N$ and $T$ should be set according to the characteristics of the specific business. For example, a large $N$ and a small $T$ are needed for the financial quantification business, which however is not necessary for relatively stable businesses such as e-commerce. In this work, we set $T$ to 5 minutes and $N$ to 96 so that occasion signals cover the time frame of last 8 hours, which is usually enough for capturing the occasion evolution process.

Combining all occasion signals in the form of sequences over time, we obtain a time series:
\begin{equation}
    \bm{S}=[\bm{S}_1, ..., \bm{S}_t, ..., \bm{S}_N] \in \mathbb{R}^{M\times N},
\end{equation}
where $\bm{S}_t=[x^o_{1t}, x^o_{2t}, ..., x^o_{Mt}]$ denotes the value vector of occasion signals at the $t$-th time step. With the time series $\bm{S}$, we aim to model the occasion evolution and obtain a well-learned occasion representation.

\section{The Proposed Method}

\begin{figure*}[t]
    \centering
    \includegraphics[width=1\linewidth]{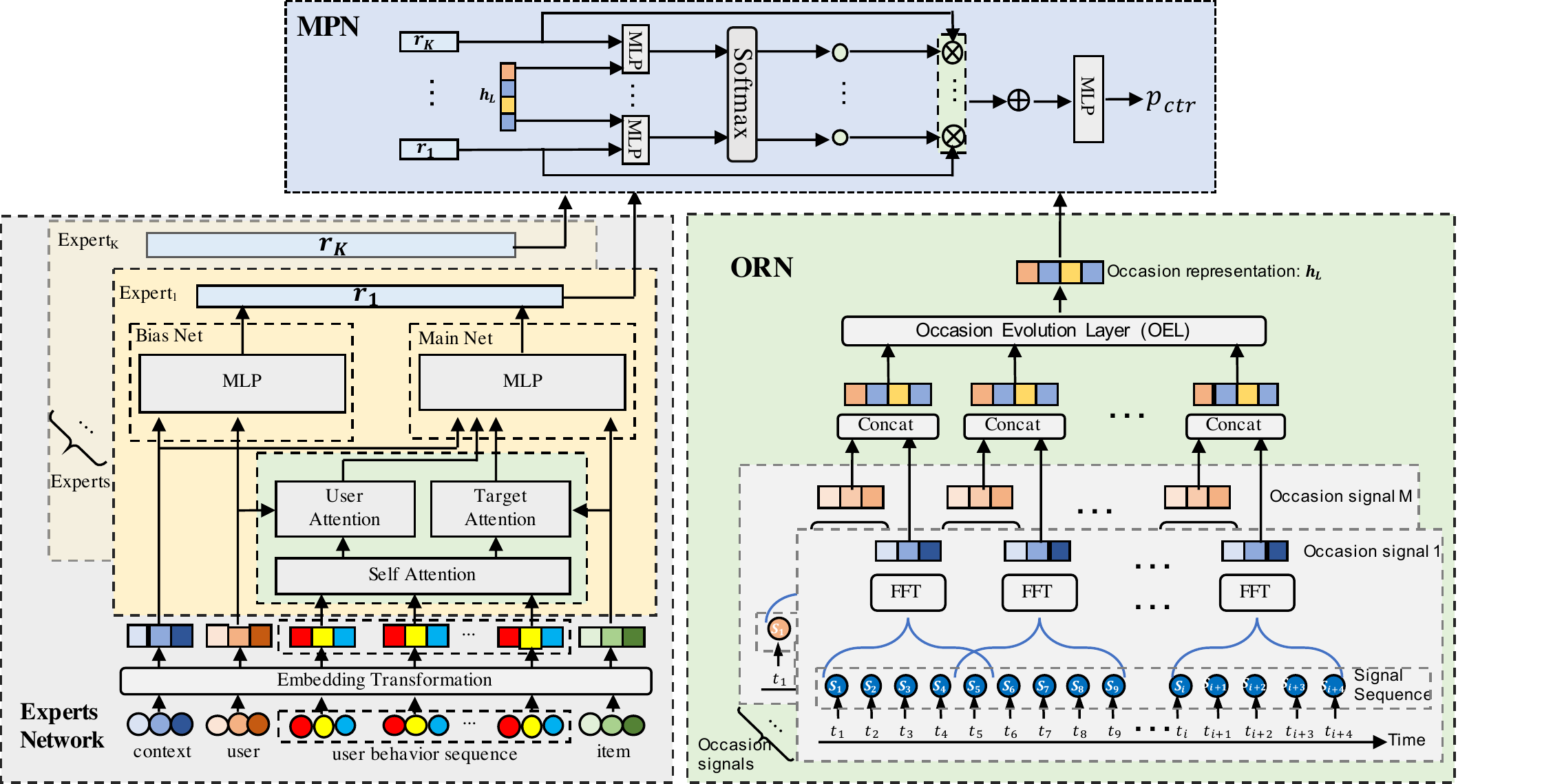}
    \caption{Framework of the proposed MOEF model, which consists of the Occasion Representation Network (ORN), Experts Network and Mixture Prediction Network (MPN). }
    \label{fig:MOEF}
\end{figure*}

In this section, we present the details of our proposed MOEF model. As shown in Figure~\ref{fig:MOEF}, occasion signals are fed into the ORN while the other features are fed into experts network. With occasion signals transformed into the frequency domain, the ORN models the occasion evolution and generates a well-learned occasion representation, perceiving the changes of occasions. Meanwhile, each expert generates a feature representation of the target item and user. With the output of the experts network and the ORN, attention weights are calculated to combine different experts for the final CTR prediction.

\subsection{Occasion Representation Network}
To model the occasion evolution, we generate sub-sequences in the time series $\bm{S}$ by applying sliding windows to $\bm{S}$ with $N_w$ as the window size and $N_s$ as the stride, therefore obtaining a sequence of sub-sequences, $i.e.$,
\begin{equation}
    \begin{split}
        &\bm{S}^s=[\bm{S}^s_1, ..., \bm{S}^s_t, ..., \bm{S}^s_L], \\
        &\bm{S}^s_t=[\bm{S}_{t\times N_s}, ..., \bm{S}_{t\times N_s+N_w-1}] \in \mathbb{R}^{M\times N_w}, \\
        &L=\lceil N/N_s \rceil,
    \end{split}
\end{equation}
where $\lceil \cdot \rceil$ refers to the round up operation and $\bm{S}^s_t$ denotes the $t$-th sub-sequence that consists of $M$ occasion signals. Generally, a larger $N_w$ and a smaller $N_s$ are more beneficial for observation over the occasion evolution. Considering both effectiveness and efficiency, we set $N_w$ to 24 and $N_s$ to 6 in this work, so each sub-sequence covers a time frame of 2 hours and the time gap between adjacent sub-sequences is half an hour.

Intuitively, with occasions changing, the frequency components of different sub-sequences are much more discriminative. Motivated by this, we perform $N_f$-point FFT on each sub-sequence to calculate its frequency components:
\begin{equation}
    \begin{split}
        \bm{F}&=[\bm{F}_1, ..., \bm{F}_t, ..., \bm{F}_L], \\
        \bm{F}_t&=|FFT(\bm{S}^s_t)| \in \mathbb{R}^{M\times N_f},
    \end{split}
\end{equation}
where $|\cdot|$ refers to the calculation of modulus of complex numbers. Then we flatten $\bm{F}_t$ so a sequence of numerical observations is obtained:
\begin{equation}
    \begin{split}
        \widetilde{\bm{F}}&=[\widetilde{\bm{F}_1}, ..., \widetilde{\bm{F}_t}, ..., \widetilde{\bm{F}_L}], \\
        \widetilde{\bm{F}_t}&=Flatten(\bm{F}_t) \in \mathbb{R}^d, \\
        d&=M\times N_f,
    \end{split}
\end{equation}
where $\widetilde{\bm{F}} \in \mathbb{R}^{d\times L}$ is a sequence of frequency spectrum, of which each position can be treated as a raw representation of the corresponding sliding window.

At last, in the Occasion Evolution Layer (OEL), we apply Long Short Term Memory (LSTM) network to $\widetilde{\bm{F}}$ to capture and characterize the global temporal dependency in the short-term sequence. The LSTM is implemented as:
\begin{equation}
    \begin{split}
        \bm{i}_t & = \sigma(\bm{W}^1_i \widetilde{\bm{F}_t} + \bm{W}^2_i \bm{h}_{t-1} + \bm{b}_i), \\
        \bm{f}_t & = \sigma(\bm{W}^1_f \widetilde{\bm{F}_t} + \bm{W}^2_f \bm{h}_{t-1} + \bm{b}_f), \\
        \bm{o}_t & = \sigma(\bm{W}^1_o \widetilde{\bm{F}_t} + \bm{W}^2_o \bm{h}_{t-1} + \bm{b}_o), \\
        \bm{c}_t & = \bm{f}_t \bm{c}_{t-1} + \bm{i}_t\,tanh(\bm{W}^1_c \widetilde{\bm{F}_t} + \bm{W}^{2}_{c} \bm{h}_{t-1} + \bm{b}_c), \\
        \bm{h}_t & = \bm{o}_t\,tanh(\bm{c}_t), \\
    \end{split}
\end{equation}
where $\bm{i}_t$, $\bm{f}_t$, $\bm{o}_t$ and $\bm{c}_t$ are the input gate, forget gate, output gate and cell vectors, respectively. The LSTM encodes the short-term sequence of frequency spectrum into a hidden output vector $\bm{h}_t \in \mathbb{R}^{d_h}$ at time $t$, where $d_h$ denotes the hidden unit size of LSTM. With $\bm{c}_t$ carrying information from $\bm{h}_{t-1}$ and flowing between cells, we can model the occasion evolution more comprehensively in a dynamic manner. The $\bm{h}_L$ is passed to subsequent neural structures as the occasion representation.

Considering the computational latency, the Transformer-Encoder \cite{vaswani2017attention}, $i.e.$, self-attention mechanism with position encoding, can be an alternative to LSTM to model the occasion evolution, especially when $L$ is large.

\subsection{Experts Network} \label{subsec:expert}
In this section, we adopt multiple experts to model the target item and user from multiple aspects, generating a set of user-item feature representations $\{\bm{r}_1, ..., \bm{r}_K\}$ for the final prediction, where $K$ denotes the number of experts.

As shown in Figure~\ref{fig:MOEF}, the input features of experts network consist of $\bm{x}^u$, $\bm{x}^i$, $\bm{x}^c$ and $\bm{x}^{seq}$, which are detailed in Section~\ref{subsec:problem}. All these features are processed by a Shared Embedding Layer so we obtain the embedded user features, item features, context features and user behavior sequence, $i.e.$, $\bm{e}^u$, $\bm{e}^i$, $\bm{e}^c$ and $\bm{e}^{seq}=\{\bm{e}^i_1,...,\bm{e}^i_m\}$, where $\bm{e}^i_m$ denotes the item embedding of $m$-th user behavior and $m$ is the sequence length.

In each expert, we perform three kinds of attention calculation. Firstly a multi-head self-attention \cite{vaswani2017attention} is calculated over $\bm{e}^{seq}$ to model user preference from multiple views of interest and $\bm{\hat{e}}^{seq}=\{\bm{\hat{e}}^i_1,...,\bm{\hat{e}}^i_m\}$ is the output. On top of the self-attention, user attention $\bm{a}^u$ is calculated to mine personalized information with $\bm{e}^{u}$ attending to $\bm{\hat{e}}^{seq}$, and target attention $\bm{a}^i$ is employed to activate historical interests related to the target item with $\bm{e}^{i}$ attending to $\bm{\hat{e}}^{seq}$. Then $\bm{a}^u$, $\bm{a}^i$, $\bm{e}^{i}$, $\bm{e}^{u}$ and $\bm{e}^{c}$ are concatenated and fed into the MainNet, $i.e.$, a Multi-Layer Perception (MLP). Meanwhile, we feed $\bm{e}^{u}$ and $\bm{e}^{c}$ into another MLP ($i.e.$, the BiasNet) to model the bias that different users in different contexts usually behave differently even to similar items. Finally, the output of an expert is obtained by concatenating the outputs of MainNet and BiasNet.

\subsection{Mixture Prediction Network} \label{subsec:MPN}
In this section, we integrate the occasion representation and experts via an attention mechanism \cite{vaswani2017attention}, $i.e.$, treating the occasion representation as Query and the output of experts as Key and Value. Thus, we can obtain the final occasion-adaptive feature representations and achieve the goal of serving promotion periods and normal days with a unified CTR model.

As shown in Figure~\ref{fig:MOEF}, the weight vector $\bm{\alpha}$ used to combine different experts is calculated via an attention mechanism:
\begin{equation}
    \alpha_i=\frac{exp(f_g(\bm{h}_L, \bm{r}_i))} {\sum_{j=1}^{K}exp(f_g(\bm{h}_L, \bm{r}_j))},
\end{equation}
where $\alpha_i$ is the weight value for the $i$-th expert and $f_g$ is a function that projects the input into a scalar. The final CTR prediction is formulated as:
\begin{equation}
    \hat{y}=f_{\theta}(\bm{x})=\mathcal{F}\left({\sum_{i=1}^{K} \alpha_i \cdot \bm{r}_i} \right),
\end{equation}
where $\mathcal{F}$ is a 3-layer MLP with a hidden unit size [144, 64, 1], of which the last layer uses Sigmoid as activation function while the other layers use ReLU. We adopt the widely-used logloss to train our MOEF model as follows:
\begin{equation} 
    L = - \frac{1}{|\mathcal{D}|} \sum_{(\bm{x},y) \in \mathcal{D}} (y\,log\,\hat{y} + (1-y)\,log(1-\hat{y})).
\end{equation}
where $\mathcal{D}$ denotes training set and $|\mathcal{D}|$ denotes the number of samples in $\mathcal{D}$.

\section{Experiments}
In this section, we conduct a series of experiments to answer the following research questions:

\noindent \textbf{RQ1} How does MOEF perform compared to representative CTR models in both normal days and promotion periods?

\noindent \textbf{RQ2} Does ORN necessarily contribute to the improvement of the performance?

\noindent \textbf{RQ3} Can ORN adapt to different occasions and guide multiple experts to learn feature representations from different aspects?

\noindent \textbf{RQ4} How does the number of experts affect the performance of MOEF?

\subsection{Experimental Setup}

\begin{table}[t] 
\centering
    \caption{Statistics of the established datasets.}
    \setlength{\tabcolsep}{2mm}{
        \begin{tabular}{c c c c c c}
            \hline
            \#Dataset & \#Users & \#Items & \#Exposures & \#Clicks & \#Purchases \\
            \hline
            $\mathcal{D}$ & 5.85M & 0.82M & 1.32B & 65.82M & 819K \\
            $\mathcal{D}_v$ & 1.87M & 0.69M & 0.18B & 9.57M & 91k \\
            \hline
        \end{tabular}
    }
    \label{tab:dataset}
\end{table}

\subsubsection{Datasets}
We establish the datasets by collecting the users' interaction logs\footnote{To the extent of our knowledge, there are no public datasets suited for this promotion-aware CTR prediction task.} from our online e-commerce platform, where promotions are highly frequent and have a considerable impact on the recommender system. In order to collect sufficient training samples, logs are sampled from 2020/10/01 to 2020/12/31, including Double 11, Black Friday, Double 12, and three member day promotions (21st of every month). The entire dataset is split into non-overlapped training set $\mathcal{D}$ (2020/10/01-2020/12/15) and validation set $\mathcal{D}_v$ (2020/12/16-2020/12/31) to avoid feature leakage. Both $\mathcal{D}$ and $\mathcal{D}_v$ cover normal days and promotion periods. Table~\ref{tab:dataset} summarizes their statistics. All the data have been anonymously processed by the log system and users' information is protected.

\subsubsection{Evaluation Metrics}
Area under ROC curve (AUC) is used as the offline evaluation metric. For online A/B testing, we choose CTR and average number of user clicks (IPV), which are widely adopted in industrial recommender systems. Improving CTR and IPV simultaneously implies not only more accurate recommendation but also more active users.

\subsubsection{Competitors}
We compare our MOEF model with two classes of the previous methods, $i.e.$, methods that aim to capture high-order feature interactions, and methods based on sequential modeling of user behaviors, which are briefly described as follows:
\begin{itemize}
    \item \textbf{DeepFM} \cite{guo2017deepfm} uses an FM layer to learn second-order feature interactions and MLP to learn high order feature interactions.
    \item \textbf{AutoFIS} \cite{liu2020autofis} identifies and selects important feature interactions for factorization models with a set of learnable architecture parameters. Note that AutoFIS use DeepFM as base model.
    \item \textbf{AIM} \cite{zhu2021aim} designs a gate-based unified framework to search proper embedding size, effective feature interactions and interaction functions. In our experiments, AIM is implemented without searching embedding size.
    \item \textbf{DIEN} \cite{zhou2019deep} models interests evolving process from user behaviors via a two-layer RNN structure with an attention mechanism.
    \item \textbf{HPMN} \cite{ren2019lifelong} adopts a hierarchical and periodical updating mechanism to capture multi-scale sequential patterns of user interests.
    \item \textbf{TIEN} \cite{li2020deep} develops a time-interval attention layer and a time-aware evolution layer to strengthen the ability to predict users' emerging interests by modeling items in a dynamic manner.
    \item \textbf{MOEF-1E} refers to MOEF that uses one expert in the Experts Network.
\end{itemize}
Input features for all the competitors listed above are the same except that DeepFM, AutoFIS and AIM discard the user behavior sequence and TIEN additionally adopts item behaviors. Except for MOEF and its variants, other models treat occasion signals as context features and process them via log1p transformation \cite{zhuang2020feature}.

\subsubsection{Implementation Details} \label{subsubsec:impl} We implement these deep learning models in distributed Tensorflow 1.4. During training, we use 3 parameter servers and 6 NVIDIA Tesla V100 16GB GPU workers. The number of points for FFT is 32 and the hidden unit size of LSTM is 96. Item ID, category ID and brand ID have an embedding size of 32 while 8 for the other categorical features. We use 8-head attention with a hidden unit size of 128 in each expert. Both MainNet and BiasNet are 3-layer MLPs, of which the hidden unit size is [480, 256, 128] and [96, 32, 16] respectively. Adagrad optimizer with a learning rate of 0.01 and a mini-batch size of 256 is used for training. The number of experts $K$ varies from 2 to 5 and is set to 2 by default. We report the results of each method under its empirically optimal hyper-parameter settings.

\begin{table}[t]
\centering
    \caption{Offline results. Bold: best. Underline: runner-up.}
    \setlength{\tabcolsep}{4.5mm}{
        \begin{tabular}{l c c}
            \hline
            \multirow{2}{*}{Model} & \multicolumn{2}{c}{AUC (mean$\pm$std.)} \\
            \cline{2-3}
            & Promotion & Normal \\
            \hline
            DeepFM & 0.7140$\pm$0.00037 & 0.7076$\pm$0.00031 \\
            AutoFIS & 0.7151$\pm$0.00036 & 0.7098$\pm$0.00029 \\
            AIM & 0.7162$\pm$0.00042 & 0.7129$\pm$0.00035 \\
            DIEN & 0.7174$\pm$0.00171 & 0.7145$\pm$0.00162 \\
            TIEN & 0.7143$\pm$0.00225 & 0.7140$\pm$0.00301 \\
            HPMN & 0.7289$\pm$0.00187 & \underline{0.7265$\pm$0.00168} \\
            MOEF-1E & \underline{0.7293$\pm$0.000103} & 0.7215$\pm$0.00089 \\
            \textbf{MOEF} & \textbf{0.7457$\pm$0.00136} & \textbf{0.7376$\pm$0.00101} \\
            \hline
        \end{tabular}
    }
    \label{tab:offline}
\end{table}

\begin{table}[t]
\centering
    \caption{Results of online A/B testing.}
    \setlength{\tabcolsep}{2.5mm}{
        \begin{tabular}{l c c c c}
            \hline
            \multirow{2}{*}{Model} & \multicolumn{2}{c}{Promotion} & \multicolumn{2}{c}{Normal} \\
            \cmidrule(r){2-3} \cmidrule(r){4-5}
            & CTR Gain & IPV Gain & CTR Gain & IPV Gain \\
            \hline
            MOEF-1E & 0\% & 0\% & 0\% & 0\% \\
            DIEN & -1.51\% & -3.32\% & -0.87\% & -1.72\% \\
            HPMN & -0.08\% & -0.13\% & +1.27\% & +2.64\% \\
            \textbf{MOEF} & \textbf{+4.23\%} & \textbf{+6.47\%} & \textbf{+4.61\%} &\textbf{+6.96\%} \\
            \hline
        \end{tabular}
    }
    \label{tab:online}
\end{table}

\subsection{Experimental Results: RQ1}
To comprehensively evaluate model performance in both normal days and promotion periods, we divide $\mathcal{D}_v$ into two parts accordingly. Each offline evaluation is repeated 3 times and experimental results are presented in Table~\ref{tab:offline}. Based on the offline evaluation, we chose several models and conducted online A/B testing lasting two weeks, and the experimental results are presented in Table~\ref{tab:online}. Note that the time frame for online A/B testing covered promotion periods and normal days and MOEF-1E was used as the baseline model. The major observations are summarized as follows:
\begin{itemize}
    \item MOEF-1E outperforms DeepFM, AutoFIS, AIM and DIEN impressively, validating the importance of modeling occasion evolution since the main difference is that MOEF-1E process occasion signals with OEL while the other models treat occasion signals as regular context features.
    \item With periodic patterns of user interests captured, HPMN becomes the runner-up method in normal days. However, during promotions when changes of occasions are more frequent, it's not comparable to MOEF-1E and MOEF since it pays little attention to the occasion evolving process and fails to adapt to different occasions. TIEN delivers unsatisfactory performance because it's hard to train the huge user embedding parameters introduced by item behaviors.
    \item For both offline and online, MOEF yields the best performance with ORN guiding multiple experts to learn feature representations from different aspects, outperforming MOEF-1E and HPMN significantly. Compared to the baseline model ($i.e.$, MOEF-1E), MOEF improves 4.23\% on CTR and 6.47\% on IPV during promotion periods while 4.61\% and 6.96\% in normal days, which helps to recommend more attractive items and improves the amount of purchases.
\end{itemize}

\begin{table}[t]
\centering
    \caption{Ablation Study.}
    \setlength{\tabcolsep}{7mm}{
        \begin{tabular}{l c}
        \hline
        Model & AUC (mean$\pm$std.) \\
        \hline
        MOEF\_w/o\_FFT & 0.7148$\pm$0.00147 \\
        MOEF\_w/o\_LSTM & 0.7304$\pm$0.00126 \\
        MOEF-TE & 0.7394$\pm$0.00118 \\
        MOEF & 0.7413$\pm$0.00122 \\
        \hline
        \end{tabular}
    }
    \label{tab:ablation}
\end{table}

\subsection{Ablation Study: RQ2}
In this section, we conduct ablation experiments to study how each component in the ORN contributes to the final performance, $i.e.$,
\begin{itemize}
    \item \textbf{MOEF\_w/o\_FFT} refers to MOEF that processes occasion signals in the time domain via log1p transformation \cite{zhuang2020feature} and then feeds them into a MLP to generate the occasion representation.
    \item \textbf{MOEF\_w/o\_LSTM} refers to MOEF that uses a MLP instead of LSTM.
    \item \textbf{MOEF-TE} refers to MOEF that uses Transformer-Encoder \cite{vaswani2017attention} ($i.e.$, self-attention mechanism with position encoding) instead of LSTM for better computational latency.
\end{itemize}
Each model is evaluated on $\mathcal{D}_v$ for 3 times and the results are detailed in Table~\ref{tab:ablation}. MOEF\_w/o\_LSTM significantly outperforms MOEF\_w/o\_FFT, validating the effectiveness of handling occasion signals in the frequency domain. Compared with MOEF\_w/o\_LSTM, MOEF further improves the performance by modeling the occasion evolution via LSTM, which helps to learn a better occasion representation in a dynamic manner. 

Moreover, we compare MOEF-TE with MOEF and observe a slight performance degradation when LSTM is replaced by Transformer-Encoder, although the latter is more efficient in computation. Finally, in our case, we decide to use LSTM since the sequence processed by the OEL is short. In cases where a long sequence is needed to model occasion evolution, we suggest using Transformer-Encoder instead.

\begin{figure*}[t]
    \begin{center}
        \subfigure[]{
        	\includegraphics[scale=0.40]{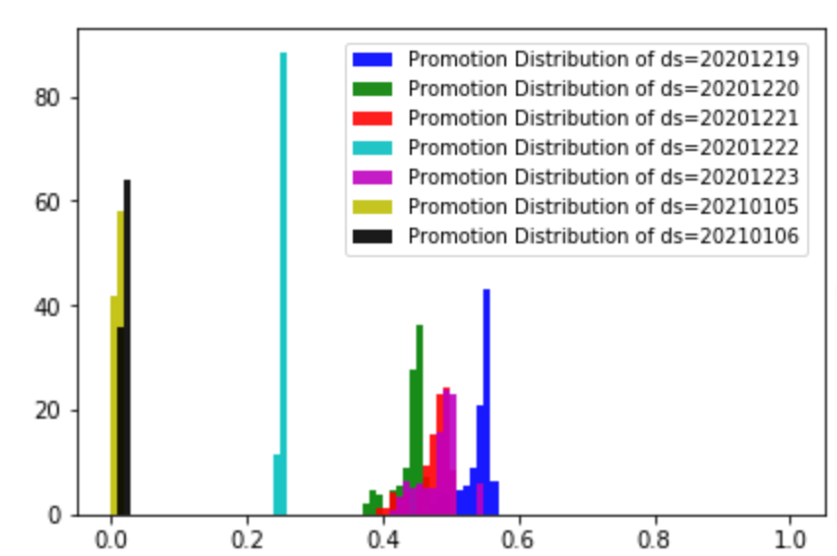}
        	\label{fig:fgn}
        }
        \subfigure[]{
        	\includegraphics[scale=0.222]{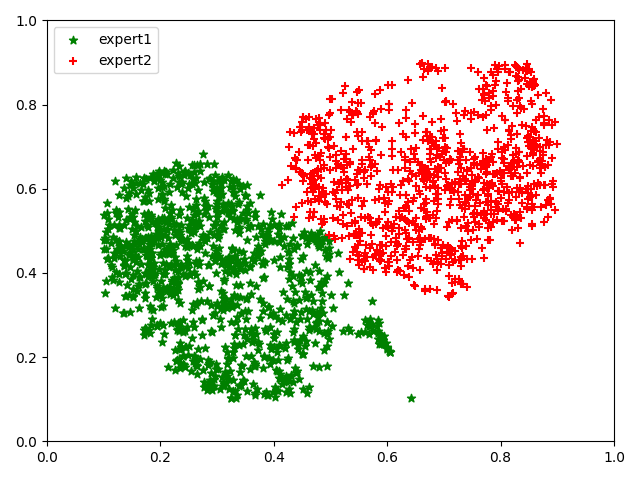}
        	\label{fig:experts_two}
        }
    \end{center}
    \vspace{-7.0mm}
    \caption{(a) Distributions of weights in different days. (b) Visualization of feature representations from two experts.}
    \label{fig:fgn_expert}
\end{figure*}

\begin{figure*}[t]
    \centering
    \includegraphics[width=1\linewidth]{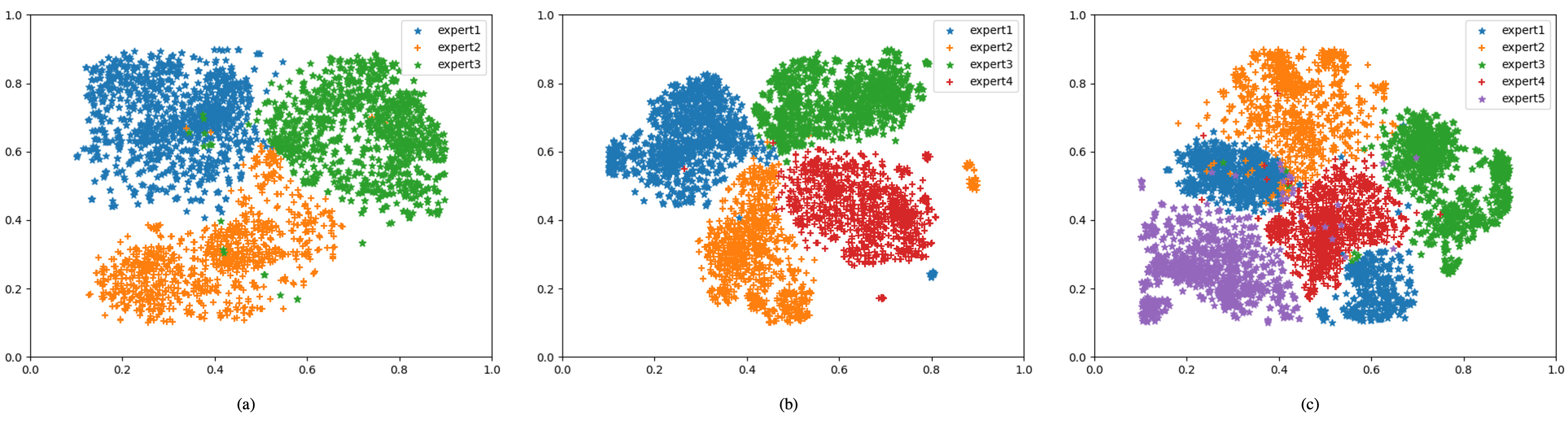}
    \caption{Visualization of feature representations from MOEF models, of which the number of experts is 3, 4, and 5, respectively.}
    \label{fig:expert_num}
\end{figure*}

\subsection{Visualization of the ORN and Experts Network: RQ3}
In this section, we conduct a visual analysis on the ORN and Experts Network to study how they contribute to the final performance. Intuitively, we attempt to observe the differences of the ORN between different occasions to validate the effectiveness of ORN, which is not easy since occasions are not explicit time periods. However, we can alternatively observe the differences of the ORN between different kinds of days of which the occasions vary greatly to study the contribution of the ORN. Specifically, we feed samples from different days into MOEF so that each sample generates a weight value $\alpha$ for one of the experts ($1-\alpha$ for another). Then, distributions of $\alpha$ in different days are plotted with different colors as shown in Figure~\ref{fig:fgn_expert}(a), of which the x-axis refers to the value of $\alpha$ and the y-axis refers to the amount of samples. It can be observed that for different kinds of days, $e.g.$, promotion periods (20201219-20201223) and normal days (20210105-20210106), the weights distinguish from each other obviously. While for the same kind of days, $e.g.$, normal days, the weights are very close. Moreover, we find that days right before promotion (20201219-20201220), middle of promotion (20201222), and the other days of promotion (20201221 and 20201223) can be further distinguished. These observations show that the ORN can perceive fine-grained occasions rather than simply distinguish promotion periods from normal days. Therefore, MOEF can adapt to different occasions and handle the distribution uncertainty.

To validate the effectiveness of the Experts Network, we randomly sample hundreds of samples and feed them into MOEF. The output of the two experts is visualized using t-SNE \cite{tsne}. As shown in Figure~\ref{fig:fgn_expert}(b), representations of the two experts clearly distinguish from each other, which implies that the Experts Network can learn feature representations from multiple aspects with multiple experts under the guidance of the ORN.

\subsection{Influence of the Number of Experts: RQ4}
In this section, we explore the influence of the number of experts by training MOEF models with the number of experts varying from 1 to 5 and evaluating each model on $\mathcal{D}_v$ for 3 times. The AUCs are 0.7263$\pm$0.00097, 0.7413$\pm$0.00122, 0.7381$\pm$0.00141, 0.7386$\pm$0.00157 and 0.7273$\pm$0.00184, respectively. Intuitively, MOEF with two experts yields better performance than that with a single expert because multiple experts can help handle the mutual interference between non-identically distributed data and learn feature representations from multiple aspects. However, with the number of experts varying from 2 to 5, we observe a decreasing trend of AUC. The visualization in Figure~\ref{fig:fgn_expert}(b) and Figure~\ref{fig:expert_num} shows that representations between experts mix with each other more seriously as the number of experts increases from 2 to 5. We suspect that the performance degradation is caused by the increased model complexity introduced by more experts, which have more learnable parameters and may lead to over-fitting.

\section{Conclusion}
In this paper, we investigate the difficulties of CTR prediction in the context of frequent e-commerce promotions and propose a novel MOEF model. Our MOEF model mainly consists of the ORN and Experts Network, where the former learns the occasion representation and guides the latter to learn good feature representations from multiple aspects using multiple experts. In this way, our MOEF model achieves promotion-aware CTR prediction and outperforms representative CTR methods on both real-world offline dataset as well as online A/B testing. The ablation study validates the effectiveness of modeling the occasion evolution in the frequency domain. Further, the visualization of weights distribution and feature representation confirms the effectiveness of our model design.

%
%
%

%




\end{document}